%% file: main.tex
\documentclass[10pt,logo,copyright]{nvidiatechreport}
\input{packages}
\input{common.tex}

\definecolor{darkred}{rgb}{0.7, 0.0, 0.0}

\usepackage[nameinlink]{cleveref}
\crefname{equation}{Eq.}{Eqs.}
\crefname{figure}{Fig.}{Figs.}
\crefname{section}{Sec.}{Sec.}
\crefname{appendix}{App.}{App.}
\crefname{table}{Tab.}{Tabs.}
\crefname{algorithm}{Algo}{Algo}
\crefname{thm}{Thm}{Thm}
\Crefname{thm}{Thm}{Thm}
\crefname{prop}{Prop}{Prop}

\newcommand{\crefnames}[3]{%
  \@for\next:=#1\do{%
    \expandafter\crefname\expandafter{\next}{#2}{#3}%
  }%
}

\title{Training Video Foundation Models with NVIDIA NeMo}

\author{NVIDIA\footnote{A detailed list of contributors and acknowledgments can be found in Section~\ref{sec:contributors} of this paper.}}

\begin{abstract}
Video Foundation Models (VFMs) have recently been used to simulate the real world to train physical AI systems and develop creative visual experiences. However, there are significant challenges in training large-scale, high quality VFMs that can generate high-quality videos. We present a scalable, open-source VFM training pipeline with \href{https://github.com/NVIDIA/NeMo/tree/main/nemo/collections/diffusion}{NVIDIA NeMo}, providing accelerated video dataset curation, multimodal dataloading, and parallelized video diffusion model training and inference. We also provide a comprehensive performance analysis highlighting best practices for efficient VFM training and inference.
\end{abstract}

\begin{document}

\maketitle

\abscontent

\input{sections/1_introduction}
\newpage % remove after adding main figure
\input{sections/2_video_curation}

\newpage
\input{sections/3_dataloading}

\newpage
\input{sections/4_training}
\input{sections/5_inference}
\input{sections/conclusion}
\clearpage
\newpage
\appendix
\input{sections/contribution}

\clearpage
\setcitestyle{numbers}
\bibliographystyle{plainnat}
\bibliography{main}

\end{document}

%% file: packages.tex
\usepackage[authoryear,sort&compress,round]{natbib}

\usepackage[utf8]{inputenc} % allow utf-8 input
\usepackage[T1]{fontenc}    % use 8-bit T1 fonts

\usepackage{parskip}        % no paragraph indents
\usepackage{url}            % simple URL typesetting
\usepackage{booktabs}       % professional-quality tables
\usepackage{amsfonts}       % blackboard math symbols
\usepackage{nicefrac}       % compact symbols for 1/2, etc.
\usepackage{microtype}      % microtypography
\usepackage{xcolor}         % colors
\usepackage{graphicx}
\usepackage{animate}        % for 360 video in teaser
\usepackage{subcaption}
\usepackage{tabularx}
\usepackage{makecell}
\usepackage{adjustbox}
\usepackage{setspace}
\newcolumntype{M}[1]{>{\centering\arraybackslash}m{#1}}
\usepackage{float}
\usepackage{tikz}
\usepackage{amsmath,amsfonts,bm, bbm,leftindex}
\usepackage{multirow}
\usetikzlibrary{arrows.meta, positioning, fit}
\usepackage{wrapfig}

%% file: common.tex
\def\eqref#1{equation~\ref{#1}}
\def\1{\bm{1}}

\DeclareMathAlphabet{\mathsfit}{\encodingdefault}{\sfdefault}{m}{sl}
\SetMathAlphabet{\mathsfit}{bold}{\encodingdefault}{\sfdefault}{bx}{n}

%% file: sections/1_introduction.tex
\section{Introduction}\label{sec::intro}
The evolution of generative AI has moved beyond text-based models with new multimodal models. These models can now handle tasks like image generation, captioning, and visual question-answering, signaling a progression towards AI systems that can interact with humans across various modalities. Recently, the focus has shifted from text and images to video, unlocking new potential applications across various industries.

Video Foundation Models (VFMs) will revolutionize several industries including robotics, autonomous vehicles, and entertainment. Building VFMs comes with unique challenges due to the scale and diversity of video data. Training VFMs requires scalable training pipelines that can ingest large amounts of data and train a model capable of understanding temporal and spatial dynamics to simulate the world. 

In this paper, we present a scalable framework for VFM training with NVIDIA NeMo. Our end-to-end training framework allows users to create or fine-tune their own VFMs. We provide high-throughput data curation with NeMo Curator, efficient multimodal data loading functionality through Megatron Energon, scalable diffusion model training leveraging Megatron Core, and a parallelized video generation pipeline in NeMo.

\begin{figure}[htbp]
\centering
\includegraphics[width=0.75\linewidth]{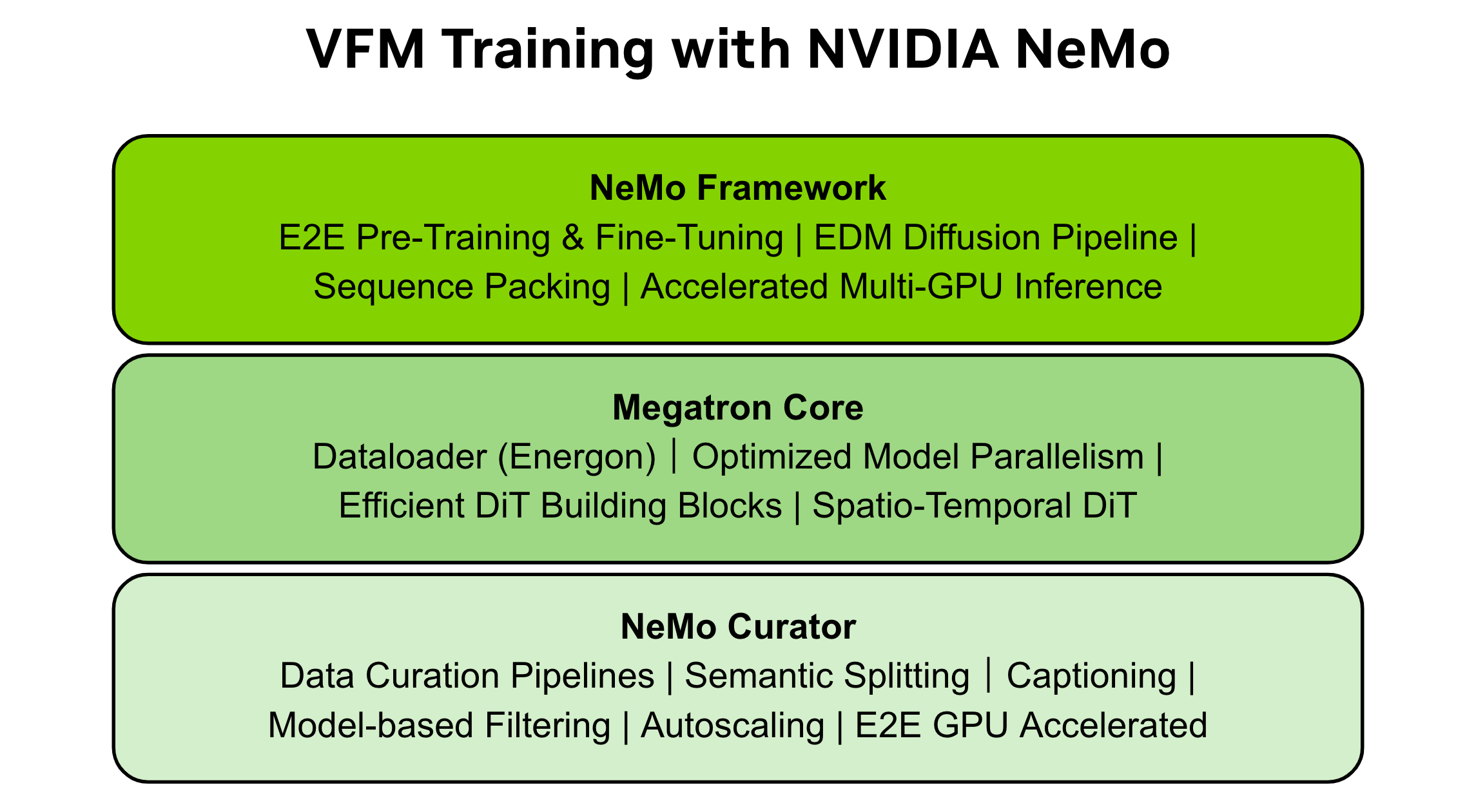}
\caption{\textbf{VFM Training Stack.} NeMo provides an end-to-end stack for training video foundation models, leveraging NeMo Curator for video curation, Megatron Core for scaling transformer models, and the NeMo Framework for pre-training, fine-tuning, and accelerated inference.}
\label{fig::main}
\end{figure}

%% file: sections/2_video_curation.tex
\section{High-Throughput Video Curation with NeMo Curator}\label{sec::curator}
Training VFMs effectively requires internet-scale high-quality video data. \href{https://developer.nvidia.com/nemo-curator}{NeMo Curator} allows users to efficiently clip, annotate and filter 100PB+ of videos. To accomplish this, NeMo Curator provides two modular pipelines that can ingest raw videos and output high quality VFM training / fine-tuning datasets. Each pipeline is composed of a sequence of stages, where each stage performs a step in the curation process. The modularity of these pipelines allows users to customize different parts of the data curation process with custom models.

\begin{figure}[htbp]
\centering
\includegraphics[width=0.99\linewidth]{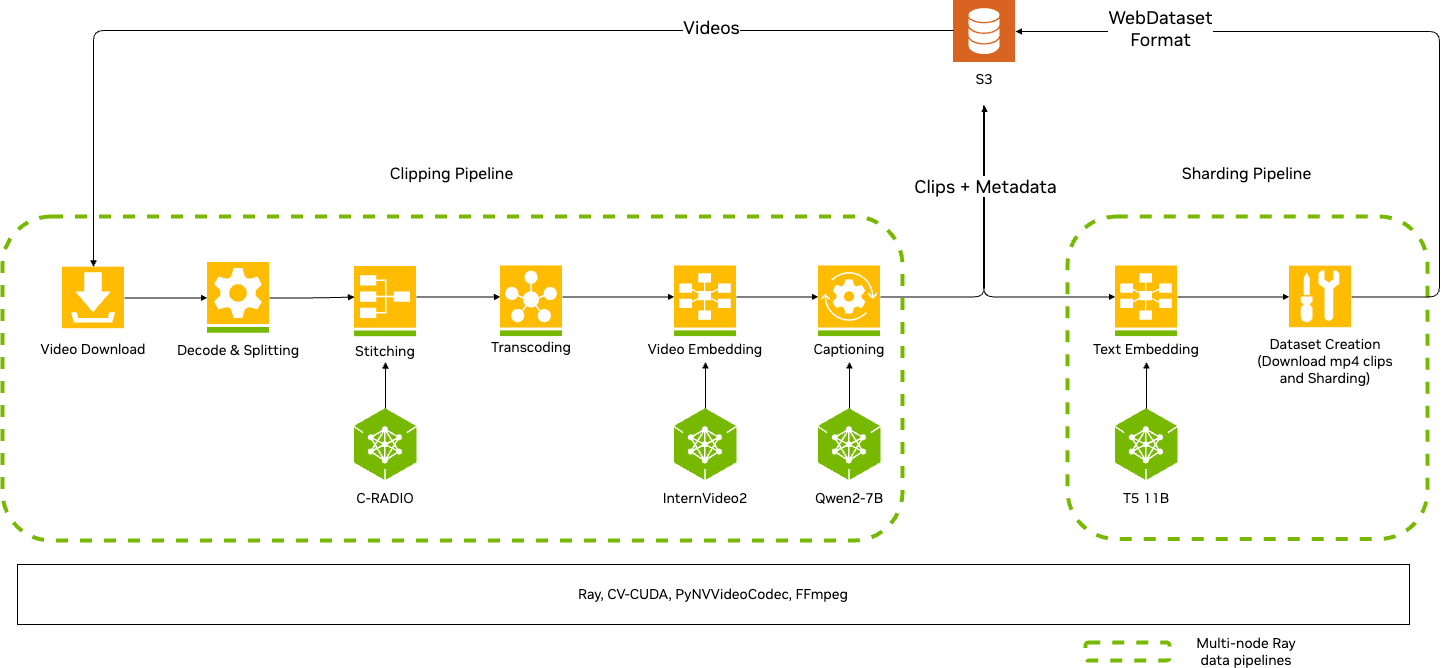}
\caption{\textbf{Video Curation Pipeline.} The video curation pipeline clips and processes large amounts of raw video. Then, the clips are sharded and stored on the cloud in the Webdataset format.}
\label{fig::nemo_curator}
\end{figure}

\subsection{Clipping Pipeline}
 
The clipping pipeline takes as input raw, uncurated videos and outputs short, continuous (no jump-cuts or other transitions) clips with associated metadata. It uses an aggressive method of splitting clips, analyzing the color changes between frames, which is smoothed out by computing the similarity between image embeddings of adjacent clips to potentially merge them back together.  These clips are then transcoded to the high-quality video encoding (H264), and they are annotated with video embeddings to facilitate semantic search capabilities. We can create synthetic captions from a VLM when no caption for the video has been provided, which has been shown to improve the quality of the downstream model over pre-written captions~\citep{sharifzadeh2024synth2boostingvisuallanguagemodels}. After this pipeline, it is recommended that users manually inspect the generated captions to verify their quality and make any alterations as they see fit.

\subsection{Sharding Pipeline}

Sharding generates text embeddings for captions and creates the final WebDataset used for training. WebDataset divides large datasets into smaller, manageable shards. This division facilitates parallel data access and processing, allowing multiple GPUs to handle different shards concurrently. This is especially necessary when training a VFM on thousands of GPUs with petabyte-scale multimodal data. The sharding pipeline splits the data into different POSIX tar archive files. During training, these tar files can be accessed with purely sequential read operations, which can signifcantly boost I/O performance from cloud storage solutions.

\subsection{Pipeline Optimizations}

The pipelines in NeMo Curator are built to scale to hundreds of nodes while running at maximum throughput. To perform effectively at scale, we use GPU acceleration in all stages of our pipelines and leverage Ray~\citep{moritz2018raydistributedframeworkemerging} to automatically scale the number of workers per stage to balance the throughput. Beyond running the model inference on GPUs, we found that utilizing the hardware video decoder (NVDEC) and hardware video encoder (NVENC) on NVIDIA GPUs brought a 3x speedup in the decoding and transcoding stages. 

\begin{figure}[htbp]
\includegraphics[width=0.99\linewidth]{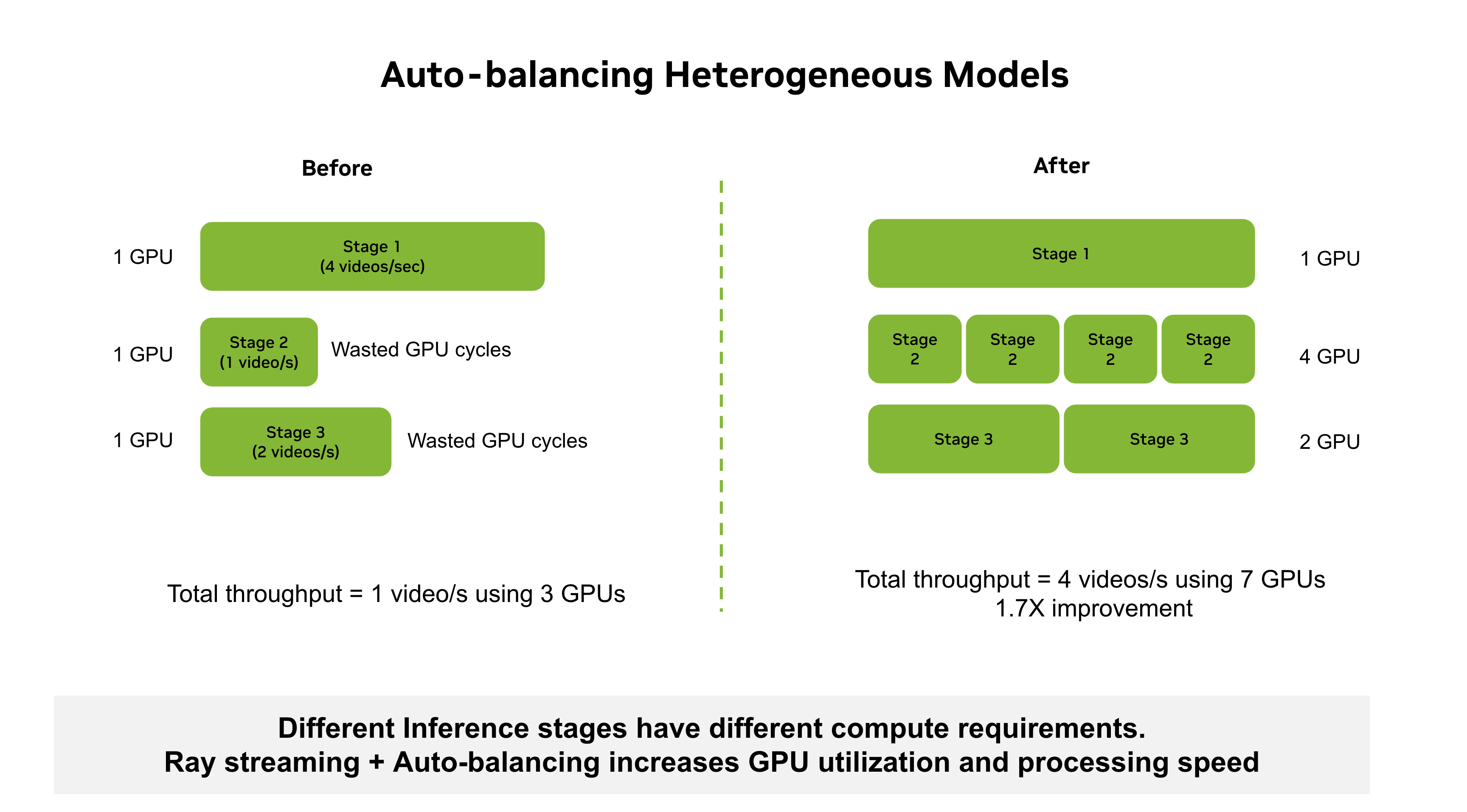}
\caption{\textbf{Auto-Balanced Curation Pipeline.} Certain curation stages can be rate-limiting the throughput of the entire curation pipeline. We created an auto-balancing system to match the throughput of the overall pipeline by allocating the optimal number of workers depending on the curation stage.}
\label{fig::auto_balance}
\end{figure}

Since each of the stages has a different throughput, it is easy to be limited by the stage with the lowest throughput. For example, in the clipping pipeline, we use different models for generating video embeddings and video captions. Video captions are generated with vision-language models that can have several billions of parameters. On the other hand, video embeddings are generated with models much smaller, around a few hundred million parameters in size. In this case, the video captioning model has much lower throughput than the video embedding model and would be a rate-limiting stage in the pipeline. Performing inference with these models without wasting GPU cycles requires a system to balance the workloads automatically. To solve this, we built an auto-balancing system as shown in Figure~\ref{fig::auto_balance} that will deploy an optimal number of workers for each stage in the pipeline. This causes  lower throughput stages to end up with more workers to keep pace with the higher throughput stages. As seen in Figure~\ref{fig::auto_balance}, we can achieve significant speedups across the overall pipeline by leveraging our auto-balancing system along with Ray for streaming distributed execution across GPU workers.

NeMo Curator integrates the clipping, sharding, and auto-balancing systems together in a single platfrom, enabling users to scale their video curation efforts. These systems are essential to training high-quality video foundation models as the quality of the model is directly dependent on the training data. NeMo Curator streamlines the process of experimenting with various data curation strategies, which accelerates the VFM research and development process.

%% file: sections/3_dataloading.tex
\section{Efficient Multimodal Dataloading with Megatron Energon}\label{sec::dataloading}
VFMs are typically trained with $\mathcal{O}$(1B) images and $\mathcal{O}$(100M) videos. An efficient dataloading strategy is necessary for high throughput during training. Our VFM training framework takes advantage of the efficient multimodal dataloading capabilities of \href{https://github.com/NVIDIA/Megatron-Energon}{Megatron Energon} to achieve this. In this section, we discuss the core Megatron Energon features that enable reliable and scalable VFM training in NeMo.

\subsection{Dataset Preparation and Blending}
Preparing and storing large-scale data for VFM training brings many challenges.
Assuming a tokenizer that compresses the temporal and spatial dimensions by a factor of 8 and fixed image/video resolution and length, one would need $\mathcal{O}$(100TB) of storage. Storing this much data on most compute clusters is typically infeasible, so we use cloud storage solutions like AWS S3 to store our training datasets. Energon uses the WebDataset format to pull dataset shards from cloud storage. Energon also enables blending multiple data sources together, allowing users to experiment and find the optimal data blend for their VFM.

\begin{figure}[htbp]
    \centering
    \includegraphics[width=0.99\linewidth]{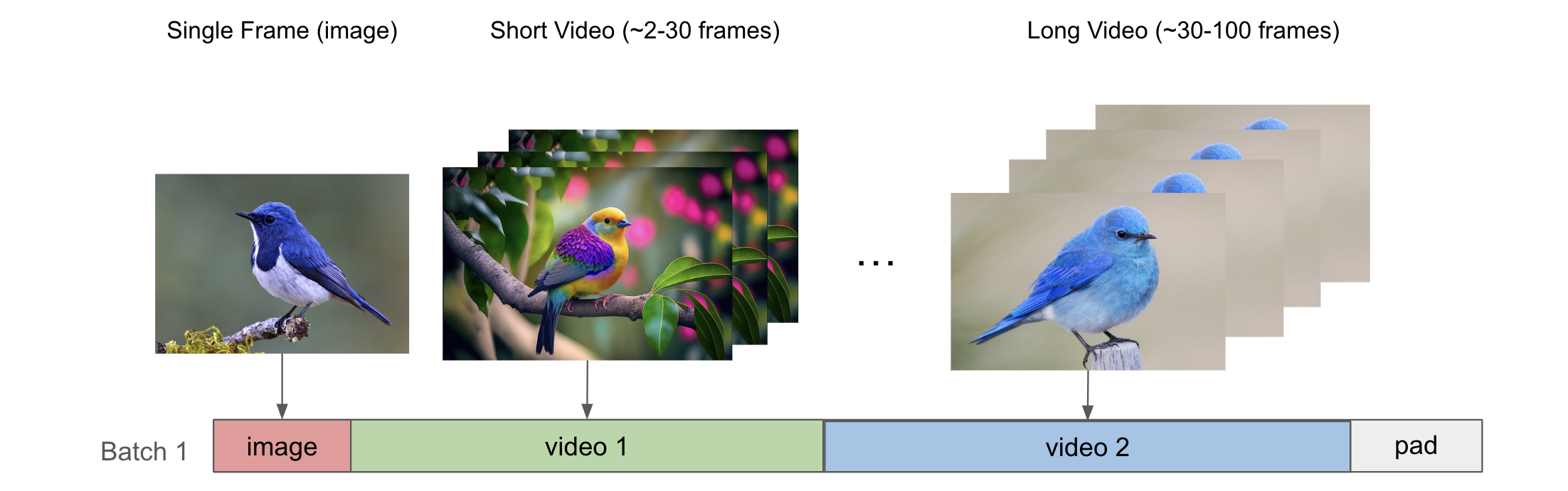}
    \caption{\textbf{Mixed Resolution Image-Video Training.} We utilize sequence packing with padding to enable joint training of images and videos with different resolutions and video length.}
    \label{fig:sequence_packing}
\end{figure}

One of the core applications of dataset blending is mixed image-video training with sequence packing. Our datasets contain both images and videos with variable sequence lengths. Typically, we can use the SBHD attention format and group sequences with the same length together in the same training batch. However, this requires that we split our video model training into different stages, introducing complex dataloading logic into our workflow. Sequence packing is a technique that concatenates multiple training samples along the sequence dimension. This method eliminates the need for excessive padding. Since more tokens are processed in each micro batch, GPU compute and GPU memory utilization are significantly increased. This is especially helpful for VFM training since it allows packing images and videos with varied length and resolution in the same micro batch as shown in Figure~\ref{fig:sequence_packing}. Although sequence packing limits the micro batch size to 1, it helps minimize compute wastage and removes the need for complicated logic, which significantly simplifies the dataloading process.  

\subsection{Optimized Network Usage}

Storing large-scale video data on cloud services like AWS S3 often strains network bandwidth during distributed training, since each model-parallel rank typically downloads the same data shards to keep their dataloaders identical. This issue intensifies as we scale to thousands of GPUs, reducing overall training throughput. Our solution is to assign a unique data shard to each rank and then use an all-gather operation to distribute these shards across ranks. This strategy maintains identical dataloaders for each rank while cutting down on redundant downloads, leading to higher training throughput in low-bandwidth environments\textemdash even with the added communication overhead. 

% Additionally, it provides accelerated sequence loading and the option to define custom data augmentations. Energon also has several features to improve reproducibility, including a data loading checkpointing system for long training runs and consistency across different training cluster setups. These abilities are crucial to enable scalable data loading for large VFM training jobs.

%% file: sections/4_training.tex
\section{Scaling Video Foundation Model Training}\label{sec::training}
NeMo provides a user-friendly interface to scale video diffusion model training. Video foundation models can be trained with autoregressive next-token prediction or full sequence diffusion objectives. NeMo’s well-established suite of tools on large language models (LLMs) can be reused for autoregressive models. In this section, we discuss our new scalable diffusion training capabilities that includes implementations of state-of-the-art diffusion transformers such as NVIDIA Cosmos diffusion world foundation models~\citep{nvidia2025cosmosworldfoundationmodel}. Our diffusion training framework provides various model parallelism options, achieving over 40\% Model FLOPs Utilization (MFU) for several model configurations.

\subsection{Diffusion Formulation}

Diffusion models learn to generate high-quality videos from random noise through an iterative sampling process. Specifically, diffusion models are trained to reverse a fixed time-dependent stochastic process that corrupts data by adding noise. We train a neural network $\epsilon_{\theta}$ that estimates the noise $\epsilon_t \sim \mathcal{N}(0, \mathbf{I})$ added to create the corrupted video latent $z_t = \alpha_t x_0 + \sigma_t \epsilon_t$. Our diffusion model can be conditioned on multiple signals, such as the noise schedule timestep $t$ and text embedding $y$. 
We train the model with the following denoising score matching loss formulation: 
$$
\mathcal{L}\left(\theta\right) = \mathbb{E}_{t \sim \mathcal{U}(1, T), \epsilon_t \sim \mathcal{N}(0, \mathbf{I})} 
\left[ w(t) \|\epsilon_t - \epsilon_\theta(z_t; t, y)\|^2 \right],
$$
where $w(t)$ is a function that weights the different denoising tasks contributing to the training objective. Additionally, we use the same denoising network preconditioning, loss weighting, and noise distribution as proposed in EDM~\citep{karras2022elucidatingdesignspacediffusionbased,karras2024analyzingimprovingtrainingdynamics}. 

We can sample videos from our diffusion model by iteratively running this denoising process for $N$ steps on random noise with a given conditioning signal $y$. Here, we can utilize various stochastic samplers to generate videos. Our framework provides both the second-order EDM Heun sampler~\citep{karras2022elucidatingdesignspacediffusionbased} the higher-order RES sampler~\cite{zhang2023improvedorderanalysisdesign} for accelerated video generation.

\subsection{Training Pipeline}
\begin{figure}[htbp]
    \centering
    \includegraphics[width=0.99\linewidth]{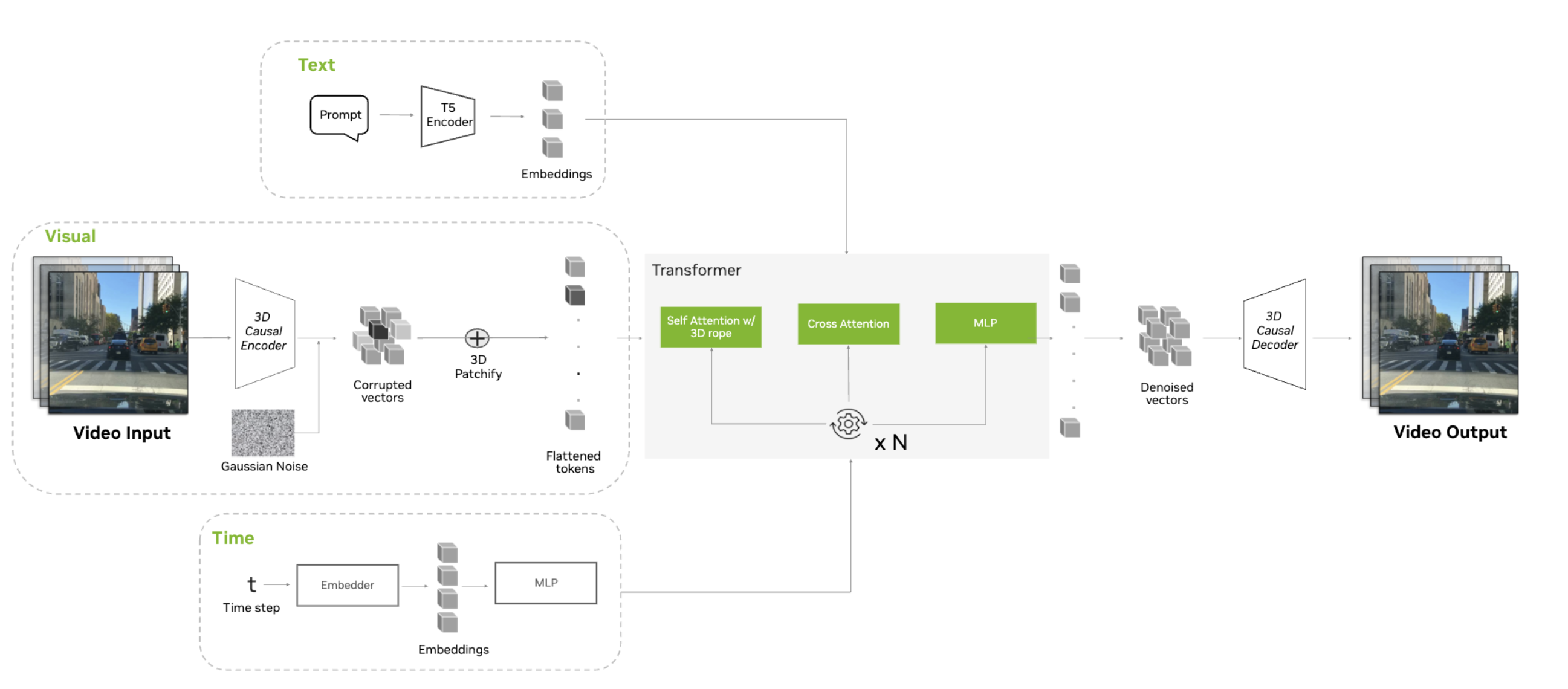}
    \caption{\textbf{Video Diffusion Transformer.} Our pipeline consists of various input signals such as text, videos, and noise timestep which are compressed and used to train a video diffusion transformer.}
    \label{fig:diffusion_training}
\end{figure}
Training video diffusion transformers requires several different components. Figure~\ref{fig:diffusion_training} displays the end-to-end process to train and generate video samples from video diffusion transformers.

The first step is video tokenization. Given an input image or video, we generate spatiotemporal tokens with a causal temporal 3D tokenizer. We flatten the 3D tokens into a sequence of 3D spacetime patches using a 3D patchify operation. We also provide video tokenizer fine-tuning capabilities in our framework. Details on tokenizer training can be found in Section~\ref{sec:appendix_vae}.

Next, we train the diffusion transformer (DiT)~\citep{Peebles2022DiT} to denoise the 3D spacetime patches that have been corrupted with Gaussian noise. The diffusion transformer will output the denoised version of the full sequence of visual tokens. During training, we condition the diffusion transformer with the diffusion noise schedule timestep $t$ and the text embedding of the prompt describing the input video sample. The timestep $t$ is applied through an Adaptive LayerNormalization (AdaLN) mechanism, in which we condition the gamma and beta parameters of the LayerNorm on $t$. One issue with using AdaLN is that a significant portion of the total model parameters consist of the MLP parameters from Adaptive LayerNorm layers. We provide the option to use AdaLN-LoRA~\citep{gupta2023photorealisticvideogenerationdiffusion}, which reduces the total parameters used by the AdaLN layers by decomposing the operation into two low-rank matrix multiplications. We observe that AdaLN-LoRA also improves MFU during training. The text conditioning is applied through a cross attention layer in each transformer block, in which we compute the attention between the video query tensor and the text embedding key-value tensors.

Finally, we compute the diffusion loss with a parallelized EDM diffusion pipeline using the noise prediction from the diffusion transformer. Our custom EDM diffusion pipeline is compatible with model parallelisms such as tensor parallel, pipeline parallel, and context parallel.

\subsection{Training a Custom Video Tokenizer}
\label{sec:appendix_vae}

Training directly on original videos or high-resolution images is often computationally prohibitive. A widely adopted approach involves encoding the original data into a latent space using a video tokenizer, which can reduce the total dataset size by over 100 times, thereby significantly accelerating the learning process. As shown in Fig.~\ref{sec::training}, the training process begins with a 3D causal video tokenizer to compress video inputs, enabling efficient VFM training. Recognizing the critical role of video tokenizers, our framework offers a comprehensive toolkit that allows users to customize or fine-tune video tokenizer models to better suit their specific needs.

\subsubsection{Customizing Tokenizer Architecture}

Our framework leverages open-source models from Hugging Face Diffusers~\citep{von-platen-etal-2022-diffusers} to enable seamless video tokenizer customization. Users can modify the video tokenizer architecture by simply adjusting a few lines in the configuration JSON files. For instance, the default video tokenizer provides an 8x spatial compression. Users can increase the spatial compression rate by adding and fine-tuning additional encoder/decoder blocks, which can significantly reduce downstream VFM training costs by further reducing total dataset size. This approach provides users with the flexibility to balance generation quality against available computational resources.

\subsubsection{Finetuning Video Tokenizers}
We support fine-tuning either pretrained tokenizers or customized tokenizers with partial weight initialization from a pretrained tokenizer on proprietary user data. Fine-tuning a pretrained tokenizer enables users to quickly adapt an existing model to their specific video data. Fine-tuning a customized tokenizer allows users to modify the video tokenizer architecture, enabling adjustments to the level of data compression based on their computational resources and downstream task requirements. We also provide users the option to use multiple loss functions, including Mean Squared Error (MSE), Kullback-Leibler (K-L) divergence loss, Learned Perceptual Image Patch Similarity (LPIPS) loss~\citep{zhang2018unreasonableeffectivenessdeepfeatures}, and Generative Adversarial Network (GAN) loss~\citep{goodfellow2014generativeadversarialnetworks}. 

\subsection{Parallelizing Diffusion Transformers}

Diffusion modeling has shown impressive scaling behaviors in terms of data, model size, and compute~\cite{li2024scalabilitydiffusionbasedtexttoimagegeneration}. The DiT architecture has shown desirable scaling properties for generative and perceptual tasks~\cite{liang2024scalinglawsdiffusiontransformers, ravishankar2024scalingpropertiesdiffusionmodels}. Modern video foundation models have adopted similar transformer-based architectures~\cite{videoworldsimulators2024,gupta2023photorealisticvideogenerationdiffusion,polyak2025moviegencastmedia,kong2025hunyuanvideosystematicframeworklarge}. However, the DiT architecture presents several challenges during both training and inference. Training DiT models with several billions of parameters on long sequences of videos has high activation memory usage, necessitating the use of various model parallelism strategies. In this section, we will discuss the various training parallelism options in NeMo enabled by \href{https://developer.nvidia.com/megatron-core}{Megatron Core} for video diffusion training.

\noindent \textbf{Tensor Parallel (TP).} Increasing the overall size of DiT models also increases the memory requirements for storing activations during training. TP splits the parameter tensors for each layer distributes them across multiple GPUs~\citep{shoeybi2020megatronlmtrainingmultibillionparameter}. Thus, each GPU holds and processes only a fraction of the total parameters. This strategy not only alleviates memory demands\textemdash since both model parameters and activations occupy less space per GPU\textemdash but it also helps accommodate extremely large models that would otherwise exceed a single GPU’s capacity. However, due to the smaller kernel workloads for each GPU, there is an increase in CPU overhead. The additional CPU overhead along with the communication cost of TP can potentially cause a drop in overall MFU.

\noindent \textbf{Fully Sharded Data Parallel (FSDP).} FSDP shards a model’s parameters and gradients and the optimizer states across multiple GPUs~\citep{zhao2023pytorchfsdpexperiencesscaling}. In a traditional data-parallel setup, each GPU maintains a full copy of the model parameters. FSDP splits model parameters into shards and stores each shard on a different GPU. As a result, no single GPU holds the entire parameter set. During the forward pass, each GPU only needs the local shards it owns, reducing memory usage. FSDP utilizes collective operations across data-parallel GPUs. For parameter computation, an all-gather is applied during both the forward and backward pass. For parameter gradients, a reduce-scatter is applied only during back-propagation. FSDP overlaps all communications with model compute, providing a seamless model parallelism experience that does not require finding an optimal configuration as when using 3D Parallelism. However, FSDP performance heavily relies on the training cluster's inter-GPU communication speed especially for large model training, where the communication overhead is quite large. This issue can be mitigated by pairing FSDP with TP. By restricting the FSDP sharding group to the data-parallel group size, TP can help reduce the model state and activation size per-GPU, which effectively reduces the communication and activation memory overhead for FSDP. 

\noindent \textbf{Context Parallel (CP).} In order to train our DiT model on long sequences of video, we must parallelize the transformer computation along the sequence dimension due to high activation memory costs. To enable long-context training, we utilize CP to parallelize the network activations across multiple GPUs by sharding the input tensors along the sequence dimension~\citep{liu2023ringattentionblockwisetransformers}. This effectively shards the activations of all layers, unlike sequence parallelism (SP), which only shards the LayerNorm and Dropout layer activations. 

When using CP, all model components besides attention are not modified since they do not have inter-token operations. For attention, we must compute each query token (Q) with the keys and values (KV) of all tokens in the sequence. CP splits Q and KV into multiple blocks, so we must ensure that the computation is done correctly by leveraging communication collectives. Specifically, we arrange the GPU hosts in a ring topology, where each GPU processes a block of Q and iteratively computes blockwise attention with each KV chunk. 

During the forward pass, an all-gather operation assembles the full KV sequence so that every Q block can attend to all tokens. In the backward pass, a reduce-scatter distributes gradients for KV back to their respective shards. To reduce activation memory cost, we only store one KV block of a sequence chunk on each GPU during the forward pass, and then we gather KV again during the backward pass. Implementing the all-gather and reduce-scatter operations as peer-to-peer (P2P) communications via TransformerEngine~\citep{nvidia_transformer_engine} allows for overlapping KV communication with blockwise attention computation, further improving training efficiency.

\noindent \textbf{Pipeline Parallel (PP).} With PP, we can shard the layers of a transformer model across multiple GPU devices, assigning an equal number of transformer blocks to each device~\citep{shoeybi2020megatronlmtrainingmultibillionparameter,narayanan2021efficientlargescalelanguagemodel}. We split a data batch into smaller microbatches, and then pipeline model execution across microbatches. One of the major challenges with PP is the pipeline bubble, which is the blocks of time where GPUs are idle. Various pipelining schedules have been proposed in recent years. In NeMo, we offer default pipelining with the GPipe algorithm~\citep{huang2019gpipeefficienttraininggiant} and interleaved pipelining~\citep{narayanan2021efficientlargescalelanguagemodel} to reduce the pipeline bubble size. 

\begin{wrapfigure}{r}{0.56\textwidth}
    \centering
    \includegraphics[width=0.56\textwidth]{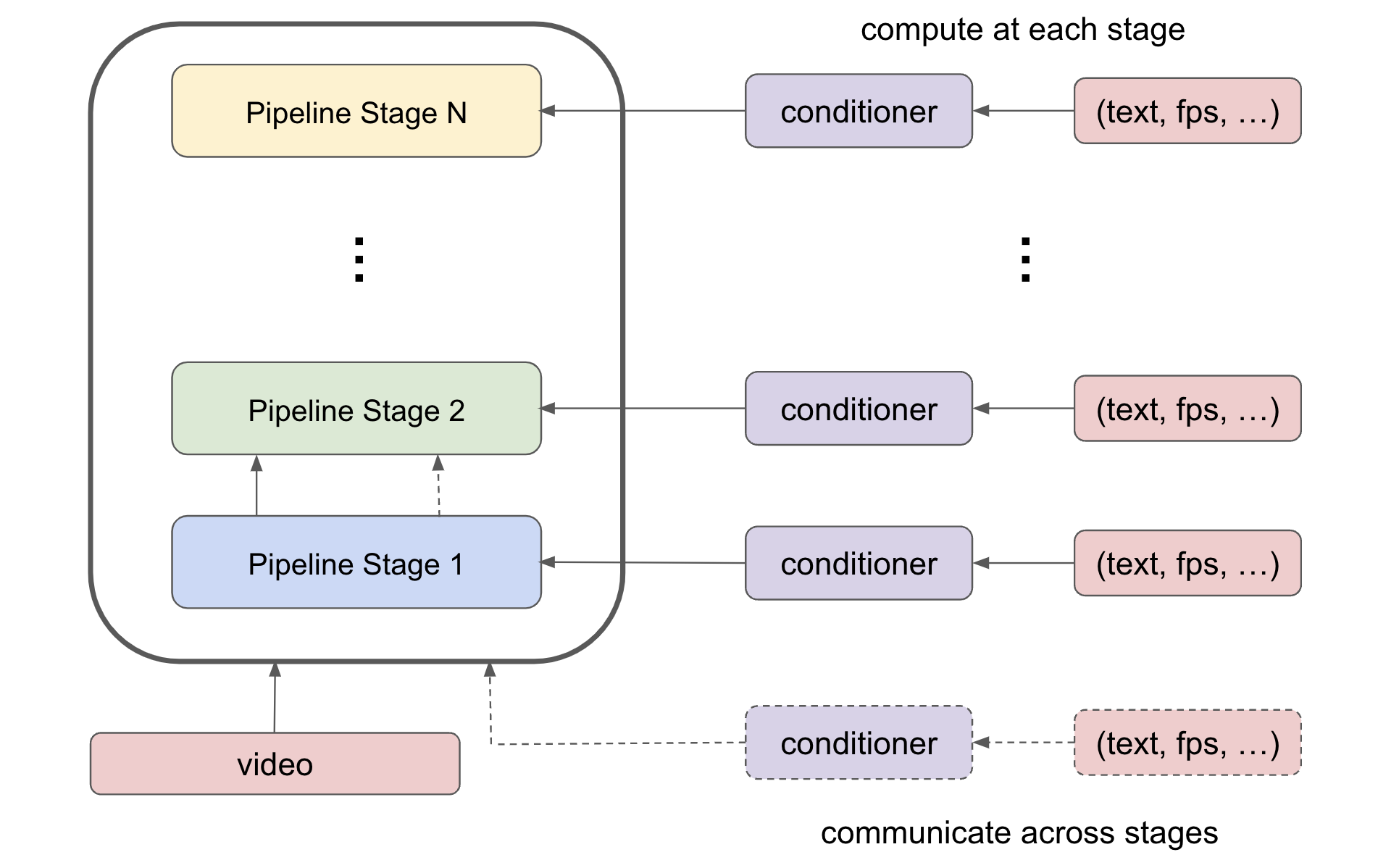}
    \caption{\textbf{Parallel Conditioning Scheme.} The DiT model uses additional conditioning signals that must be available at each pipeline stage. We determined that recomputing these embeddings at each pipeline stage is more efficient than communicating them across pipeline stages during training.}
    \label{fig:pipeline_conditioning}
\end{wrapfigure}

For the DiT model, we specifically explore how to efficiently pipeline conditioning information for the model across pipeline stages. PP for transformer models typically only communicates the hidden states across different PP stages. However, for DiT-based video foundation models, we may have several additional conditioning signals, including text embeddings, image embeddings, camera parameter embeddings etc. These embeddings are typically used in cross-attention blocks throughout each transformer block. To ensure correct training, we must be able to compute cross-attention with these conditioning signals at each pipeline stage. 

We consider two approaches to mitigate this issue as shown in Figure~\ref{fig:pipeline_conditioning}. The first method is communicating the additional conditioning signals along with the hidden states across pipeline stages. This requires computing the embeddings once before the first DiT block and then communicating them across each PP stage. The second method is computing each embedding at each PP stage separately, and then conditioning the model with the computed embeddings. In the first method, we must modify the PP schedule each time we incorporate a new conditioning signal by increasing the communication buffer size to add the new embedding, which increases communication cost. In the second method, we do not modify the PP schedule, but we must spend more compute since we recompute conditioning signals at each PP stage. We observed that the additional computation cost of the second method has a better tradeoff with MFU than the first method with the additional communication cost. It is also simpler to implement and easily scales as new conditioning signals are added.

\subsection{DiT Parallelization Performance Study}

This section provides detailed evaluation and performance benchmarks for our framework. We first explore the algorithm-system co-design space to find the best performing model configurations. This is followed by detailed performance benchmarking results, comparison with other open-source alternatives like Fast-DiT~\citep{fang2024xdit} and scaling experiments. We end this section highlighting the importance of a flexible training framework for diffusion models and a brief summary of our learnings.

\subsubsection{Algorithm-system co-design.} To achieve high compute efficiency while scaling both model sizes and context lengths, we explore algorithm-system co-design space along with parallelization strategies. 

\textbf{AdaLN-LoRA architecture}. One of the key changes we make to the DiT Architecture is altering the AdaLN projection. The AdaLN projection consumes a vast majority of the model’s parameters \verb|[Hidden Dim x 9*Hidden Dim]|. For example, using a base DiT model architecture, a 7B model would have an AdaLN projection that alone would consume 2.7B parameters.

\begin{figure}[htbp]
    \centering
    \includegraphics[width=0.55\linewidth]{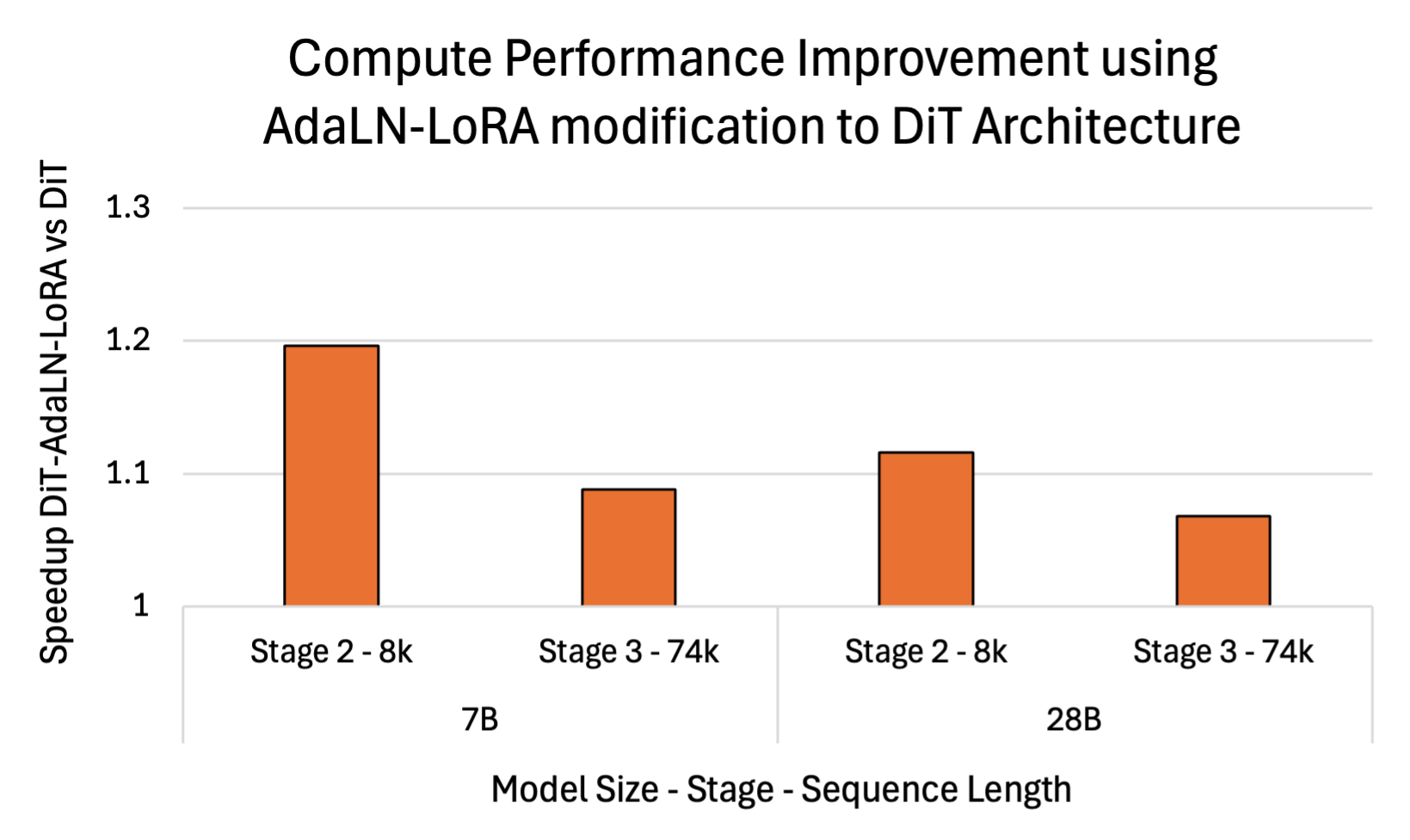}
    \caption{\textbf{Compute Performance Improvement with AdaLN-LoRA.} We benchmark the DiT training speedup with AdaLN-LoRA for 7B and 28B models on sequence lengths of 8k and 74k tokens. We observe a noticeable speedup across all configurations and up to 1.2x for DiT-7B at 8k sequence length.}
    \label{fig:dit_adaln_improvement}
\end{figure}

Additionally, given that the AdaLN is performed on the timestep embedding (instead of the full context length) and is a singular GEMM, it also doesn’t allow for a straightforward model parallelism strategy that would allow for balancing the compute with communication. To improve this, we implement AdaLN-LoRA~\citep{gupta2023photorealisticvideogenerationdiffusion}. This not only allows us to reduce the parameter density focused in AdaLN but also allows us to use larger hidden dimensions for a given fixed model size. This results in more efficient GPU compute utilization and better compute-communication ratios which helps us improve our compute performance by up to 1.2x as shown in Figure~\ref{fig:dit_adaln_improvement}.

\textbf{Modularized Context Parallelism}. Another choice we make while exploring parallelization strategies is to disable context parallelism for the CrossAttention portion of the layer when CP is used. We make this decision as the KV context length is not sufficiently large to require parallelism or overlap communication under compute if parallelism was used.

\textbf{QK Normalization}. Another challenge we face is the normalization of $Q$ and $K$ before dot product attention compute. Under Tensor Parallelism, $Q$ and $K$ are normally sharded across the number of attention heads and computing normalization across the entire hidden dimension would require excessive communication. We choose to limit this normalization per head in order to improve compute throughput and find no adverse effect on model performance.

\subsubsection{Performance Benchmarking}

We study the compute throughput performance of training workloads with varying model sizes, and context lengths with multiple parallelization strategies and compute scales and show that \textbf{our training framework can achieve up to 48.2\% MFU}. 

\textbf{Performance Evaluation}. Table~\ref{table::perf_workloads} describes the workload configurations used. We report best achieved performance on 8 8xH100 nodes for all workloads except 28B - Stage 3 which is reported on 32 8xH100 nodes. 

\begin{table}[htbp]
    \centering
    \caption{\centering DiT Model Performance Benchmarking Configurations}
    \resizebox{0.6\linewidth}{!}{
    \begin{tabular}{@{}ccccc@{}}
        \toprule
        Workload & Layers & Hidden Size & Heads & Context Length \\
        \midrule
        7B -- Stage 2  & 28 & 4096 & 32 & 8192 \\
        7B -- Stage 3  & 28 & 4096 & 32 & 73728 \\
        28B -- Stage 2 & 48 & 6144 & 48 & 8192 \\
        28B -- Stage 3 & 48 & 6144 & 48 & 73728 \\
        \bottomrule
    \end{tabular}
    }
    \label{table::perf_workloads}
\end{table}

\begin{figure}[htbp]
    \centering
    \includegraphics[width=0.55\linewidth]{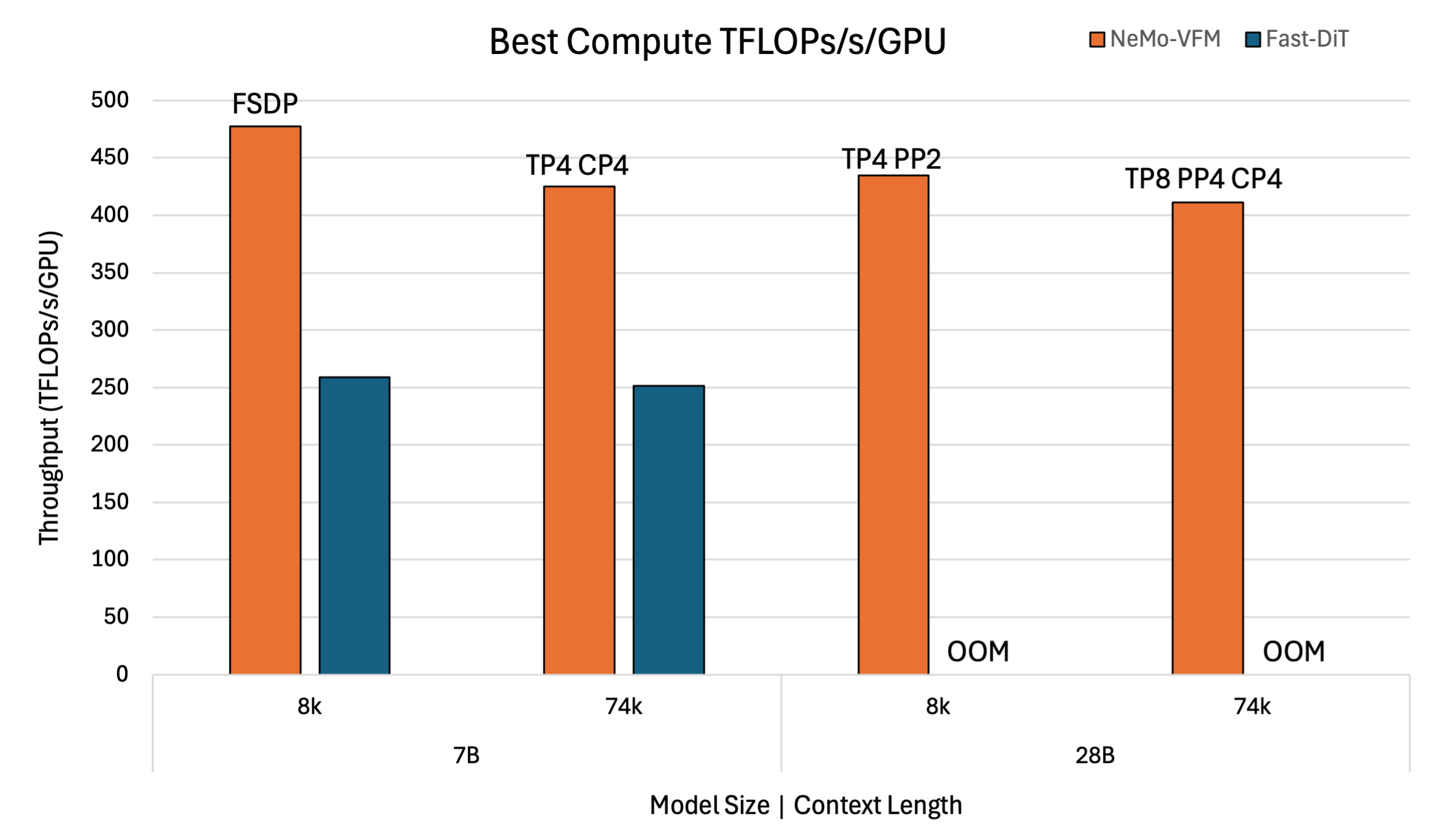}
    \caption{\textbf{Best Compute Performance per Workload.} We benchmark our training framework and compare it with the Fast-DiT framework. Our framework consistently delivers higher training throughput than Fast-DiT, and it also supports training larger models that Fast-DiT cannot handle.}
    \label{fig:fast_dit_comp}
\end{figure}

We take Fast-DiT~\citep{jin2024fast} as a baseline to compare with our framework. For fair comparison, we enable AdaLN-LoRA and FSDP support using Huggingface Accelerate~\citep{accelerate} in Fast-DiT. Figure~\ref{fig:fast_dit_comp} shows the best compute performance for each workload using our framework and compares it to the best compute performance achieved using Fast-DiT. Our training framework is able to achieve high throughput consistently across all workloads and outperforms Fast-DiT by up to 1.85x on the 7B model. Our framework maintains strong performance when scaling the model size to 28B, while Fast-DiT is unable to run the 28B model as it runs out of memory capacity.

\begin{figure}[htbp]
    \centering
    \includegraphics[width=0.5\linewidth]{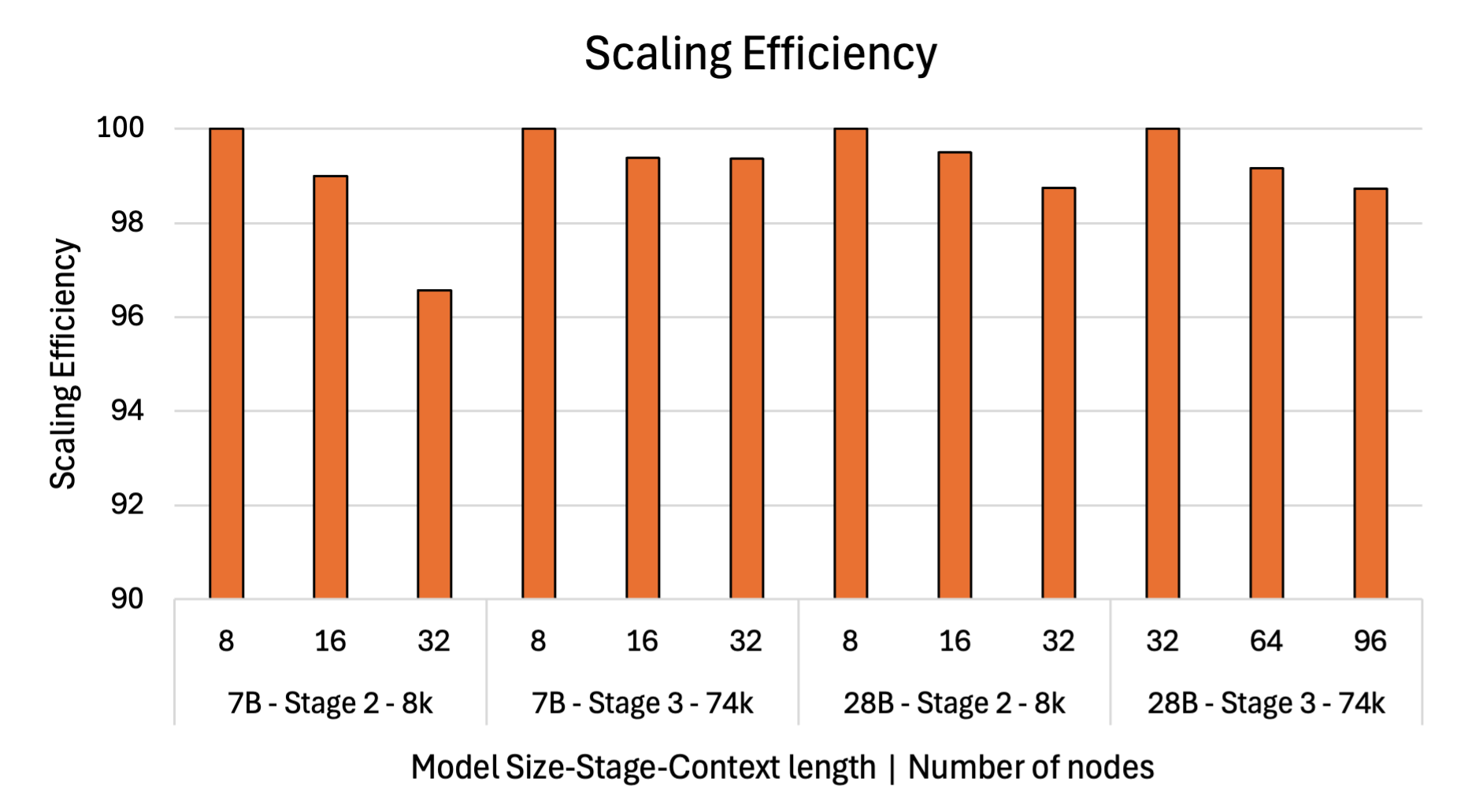}
    \caption{\textbf{DiT Training Scaling Efficiency.} We measure the scaling efficiency across four workloads, scaling from 8 to 32 nodes of 8xH100 GPUs. All configurations exceed 95\% efficiency, with most surpassing 98\%. These results highlight our framework’s near-linear scaling across training workloads.}
    \label{fig:scaling_efficiency}
\end{figure}

\textbf{Scaling Efficiency.} Our  framework also achieves high strong scaling efficiency. We continue to measure compute throughput while scaling up the number of 8xH100 nodes given a fixed workload and find that our framework achieves ~95\% or higher strong scaling efficiency as shown in Figure~\ref{fig:scaling_efficiency}.

\begin{figure}[htbp]
    \centering
    \includegraphics[width=0.7\linewidth]{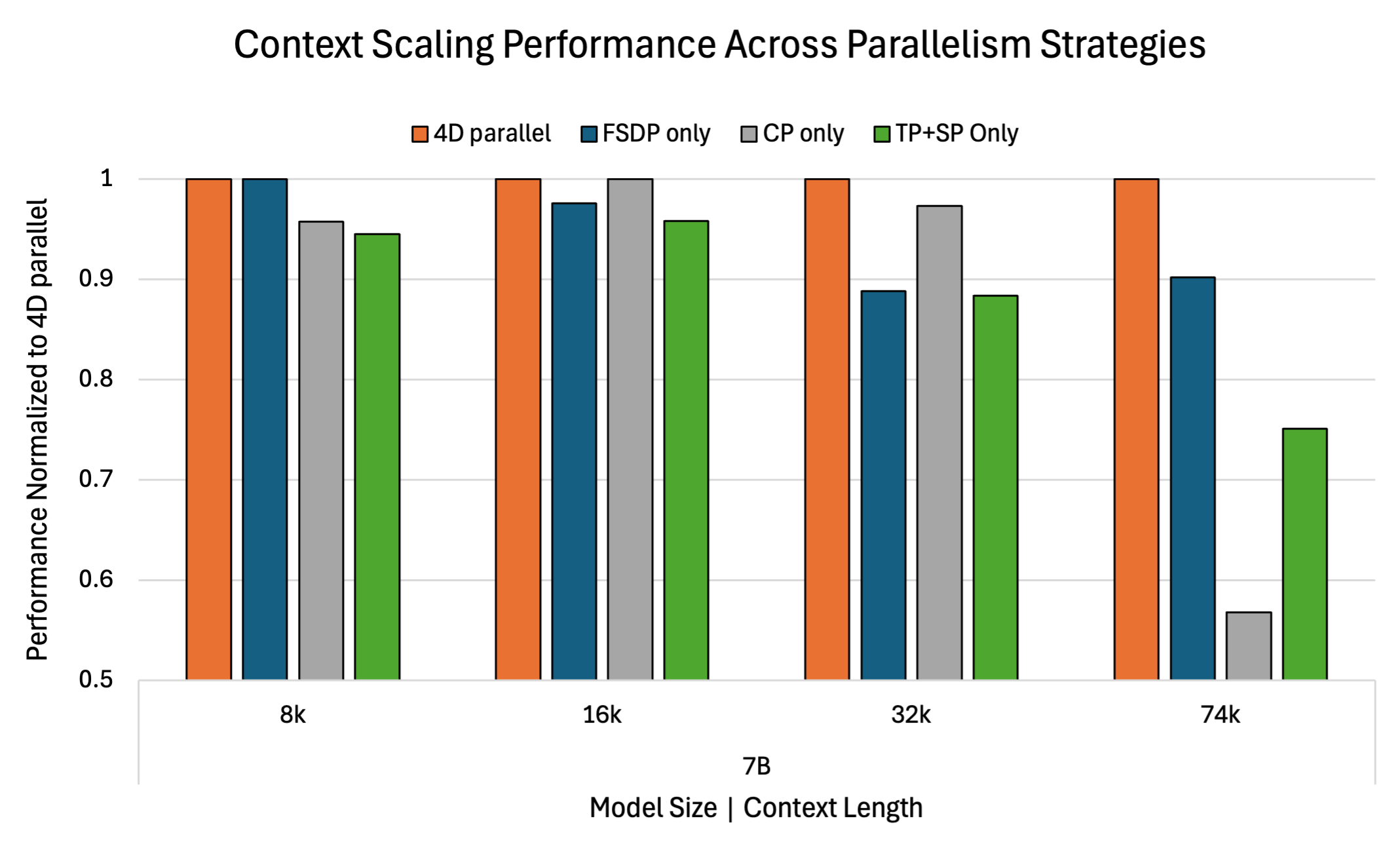}
    \caption{\textbf{Context Scaling Performance Across Parallelism Strategies.} We observe that the optimal parallelism configuration changes based on the context length. FSDP is performant for shorter context lengths, but 4D parallelism becomes necessary for efficient long-context training.}
    \label{fig:4d_parallel}
\end{figure}

\textbf{Observations and Learnings}. As shown in Figure~\ref{fig:fast_dit_comp}, it is clear that each workload requires a unique combination of parallelism strategies to achieve high efficiency, highlighting the need for a flexible training framework. To further underscore the importance of full 4D parallelism (TP, CP, PP and FSDP), we study the best throughput achieved by DiT-7B with AdaLN-LoRA under various parallelism configurations while scaling context length. Figure~\ref{fig:4d_parallel} shows performance of different strategies normalized to the best performing 4D parallel strategy. 

The best performing parallelism strategy is unique at each context length, and no single parallelism strategy alone is sufficient to consistently achieve good performance. At shorter context lengths and smaller model sizes, FSDP is typically sufficient to achieve decent training throughput. However, as the context length and activation memory footprint grows, FSDP alone is no longer the best performing configuration and requires a combination of TP, CP and FSDP. PP becomes a necessary strategy as well to distribute large models while maintaining high compute efficiency. We briefly summarize our learnings below.

\noindent \textbf{Tradeoffs Between Parallelism Strategies.} When scaling diffusion transformer training, it is important to consider the tradeoffs between specific types of parallelism strategies. TP, CP, and PP have various advantages and disadvantages when applied to the DiT architecture:
\begin{enumerate}
    \item Context Parallelism: CP primarily reduces the memory footprint of activations on each device during training. In scenarios with long video sequences, we can effectively hide the communication overhead of CP. However, when sequences are short, this overhead becomes largely exposed.
    \item Tensor Parallelism + Sequence Parallelism: TP and SP together effectively minimize the memory usage of parameters and activations. However, due to the widespread use of Adaptive Layer Normalization (AdaLN) in the DiT architecture, TP communication cannot be well overlapped with computation. Additionally, when model size is not large (i.e. $\leq$ 10B parameters), excessive splitting of parameters can lead to reduced efficiency in matrix multiplications.
    \item Pipeline Parallelism: While PP does not help in reducing activation memory usage and introduces bubbles of idle GPU time at the start and end of processing, it can effectively decrease parameter memory usage. Therefore, it is beneficial when dealing with larger model sizes, especially in settings with low communication bandwidth between GPUs.
\end{enumerate}
In summary, after weighing the advantages and disadvantages of the parallel techniques, the following recommendations can be made:
\begin{enumerate}
    \item When both the model size and context lengths are relatively small, FSDP can be sufficient.
    \item When the model size is relatively small and the video sequences are large, CP should be prioritized.
    \item As context length grows, CP should be prioritized, and as model size grows, TP/FSDP should be prioritized for any model sharding.
    \item If model size is large, TP should be prioritized intra-node to an extent where smaller GEMMs are still efficient. Beyond that, PP should be used.
    \item In cases where the model size and context length are particularly large, a combination of TP, PP and CP may be required. As a general rule of thumb, TP/SP communications should be kept intra-node while CP and PP groups can perform inter-node communication while still sufficiently hiding the communication overhead.

\end{enumerate}

\subsection{Spatial-Temporal DiT (ST-DiT)}

In addition to full attention, ST-DiT (Spatial-Temporal DiT)~\citep{opensora} introduces spatial and temporal attentions, which are applied to intra-frame and inter-frame tokens respectively. These attentions are stacked in a transformer layer, and their computational complexity varies a lot for their substantially different batch sizes and sequence lengths. 

Full attention flattens all tokens of all frames into a sequence, which results in small batch size (e.g., 1 or 2) and long sequence length (e.g., 256K, 512K, etc.). Spatial attention is applied to each frame separately which translates to a relatively large batch size (e.g., hundreds of frames) but short sequence length (e.g., a few thousands of tokens in each frame). Temporal attention is like spatial attention, because their batch sizes and sequence lengths are transposed to each other. Input shape variance leads to different tradeoffs on parallelization strategies: larger batch sizes prefer DP (data parallelism), and longer sequence lengths need CP (context parallelism). 

Since DP splits the batch dimension, it does not trigger any inter-GPU communication for attention. Thus, each GPU can work independently with its assigned batch split. CP splits the long sequence into chunks and distributes them between GPUs, so each GPU only needs to compute and save activations of a sequence chunk. However, CP inserts an all-gather to collect the full sequence of KV, because Q of each token needs to compute with the KV of all tokens in the same sequence. The CP implementation in TransformerEngine changes the all-gather to P2P communications arranged in a ring-topology, which hides the communication overhead under the attention compute if the sequence is long enough. 

\begin{wrapfigure}{r}{0.605\textwidth}
    \centering
    \includegraphics[width=1\linewidth]{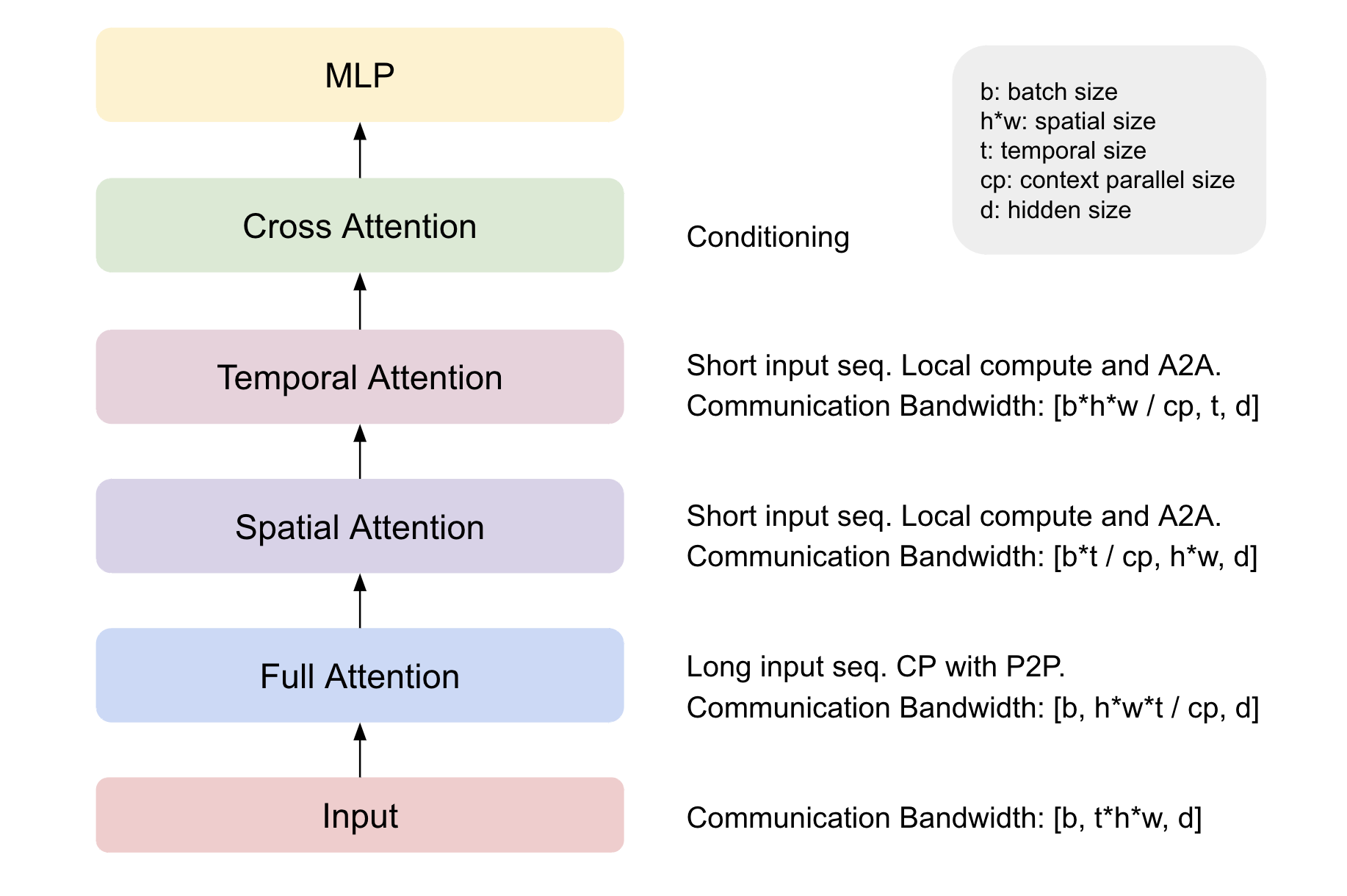}
    \caption{\textbf{ST-DiT Hybrid Parallelism.} We propose a new hybrid parallelism approach for ST-DiT models that reduces the total necessary all-to-all (A2A) collectives while fully hiding the communication overhead of full attention and minimizing the communication overhead for both spatial and temporal attention.}
    \label{fig:st_dit}
\end{wrapfigure}

We propose to apply CP to full attention, but DP to spatial and temporal attentions as shown in Figure~\ref{fig:st_dit}. This cannot work out directly due to the mismatched input/output shapes. Assume that the input shape of each transformer layer is \verb|[b, h*w*t, d]|, where $b$ is the batch size, $h$ and $w$ are the height and width of each frame, $t$ is the number of frames, and $d$ is the hidden size. Sequence length for each attention is determined by $h$, $w$, and $t$. The output shape of full attention with CP is \verb|[b, (h*w*t)/CP, d]|, but the following spatial attention requires input shape of \verb|[(b*t)/CP, h*w, d]|, and the last temporal attention expects input shape of \verb|[(b*h*w)/CP, t, d]|. To get the expected input shape for each attention, we insert all-to-all communications between attentions for tensor shape transitions.

Existing ST-DiT implementations either did not consider all three attentions simultaneously, i.e. DSP~\citep{zhao2024dspdynamicsequenceparallelism}, or had significant communication overhead exposed, i.e. 4 all-to-all collectives per attention in DeepSpeed Ulysses~\citep{jacobs2023deepspeedulyssesoptimizationsenabling}. Compared to prior implementations, our proposal not only takes all attention modules into consideration, but also can fully hide communication overheads of full attention and introduces the minimum overheads to spatial and temporal attentions (only two all-to-all collectives). 
\iffalse
    In Table~\ref{table:stdit_performance}, we display ST-DiT performance results for various model parallelism and context length configurations. Our implementation is able to achieve up to 40\% MFU on NVIDIA H100 GPUs for our 12B ST-DiT configuration.

\begin{table}[htbp]
    \centering
    \caption{\centering ST-DiT Benchmark Results}
    \resizebox{\linewidth}{!}{
    \begin{tabular}{@{}ccccc@{}}
        \toprule
        Model Size & Context Length & Parallelism Config & GPU MFU (TFLOPS/s) & Throughput (token/s/GPU) \\ \midrule
        STDiT-7B  & 35K            & TP=2, SP, PP=4, VPP=2, CP=4   & 318.7                & 7991                     \\
        STDiT-12B & 35K            & FSDP, CP=8                  & 402.5                & 5760                     \\
        STDiT-12B & 35K            & TP=2, SP, PP=4, CP=4          & 385.0                & 5485                     \\
        STDiT-7B  & 74K            & TP=2, SP, PP=2, CP=8          & 335.6                & 8124                     \\
        STDiT-12B & 74K            & TP=2, SP, PP=4, CP=8          & 381.1                & 5285                     \\ \bottomrule
    \end{tabular}
    }
    \label{table:stdit_performance}
\end{table}
\fi
Table~\ref{table:stdit_performance2} compares the performance of different parallel strategies for spatial-temporal attention. It can be seen that using all-to-all for spatial-temporal attention achieves better performance than CP, and it is able to achieve up to 40\% MFU on NVIDIA H100 GPUs for our 12B ST-DiT configuration.

\begin{table}[h]
    \centering
    \caption{\centering ST-DiT Benchmark Results}
    \resizebox{\linewidth}{!}{
    \begin{tabular}{@{}cccccc@{}}
        \toprule
        \textbf{Model Size} & \textbf{Context Length} & \textbf{Training Config} & \thead{\textbf{GPU MFU} \\   (TFLOPS/s)} & \thead{\textbf{Throughput} \\ (token/s/GPU)} & \textbf{Speedup} \\ 
        \midrule
        STDiT-7B & 35K & TP=2 SP PP=4 VPP=2 CP=4 & 284.99 & 7138.9 & - \\ 
        STDiT-12B & 35K & FSDP CP=8 & 327.61 & 4670.7 & - \\ 
        STDiT-12B & 35K & TP=2 SP PP=4 CP=4 & 359.92 & 5131.6 & - \\ 
        STDiT-7B & 74K & TP=2 SP PP=2 CP=8 & 139.4 & 3371.4 & - \\ 
        STDiT-12B & 74K & TP=2 SP PP=4 CP=8 & 168.8 & 2342.8 & - \\ 
        STDiT-7B & 35K & TP=2 SP PP=4 VPP=2 All2All=4 & 318.7 & 7984.5 & 1.118 \\ 
        STDiT-12B & 35K & FSDP All2All=8 & 402.5 & 5738.5 & 1.229 \\ 
        STDiT-12B & 35K & TP=2 SP PP=4 All2All=4 & 385.0 & 5489.2 & 1.070 \\ 
        STDiT-7B & 74K & TP=2 SP PP=2 All2All=8 & 335.6 & 8113.6 & 2.407 \\ 
        STDiT-12B & 74K & TP=2 SP PP=4 All2All=8 & 381.1 & 5288.5 & 2.257 \\ 
        \hline
    \end{tabular}
    }
    \label{table:stdit_performance2}
\end{table}

%% file: sections/5_inference.tex
\section{Efficient Video Generation Inference}\label{sec::inference}
Video diffusion models must denoise long sequences of 3D spatiotemporal tokens during inference time. To generate high resolution video samples, these models must denoise tens of thousands of tokens, which may only amount to a few seconds of actual video. The sampling process is iterative, which can lead to long generation times per sampled video.

We introduce an efficient approach for video generation using diffusion models that leverages context parallelization across multiple GPUs. In the following sections, we detail our inference approach and performance results across a variety of configurations.

\subsection{Parallelized Inference Algorithm}

We apply a simple, yet effective approach to parallelize video diffusion model inference with context parallelism. This method significantly accelerates the inference process by distributing the denoising workload across GPUs and denoising sequence chunks in parallel.

\begin{figure}[htbp]
    \centering
    \includegraphics[width=0.5\linewidth]{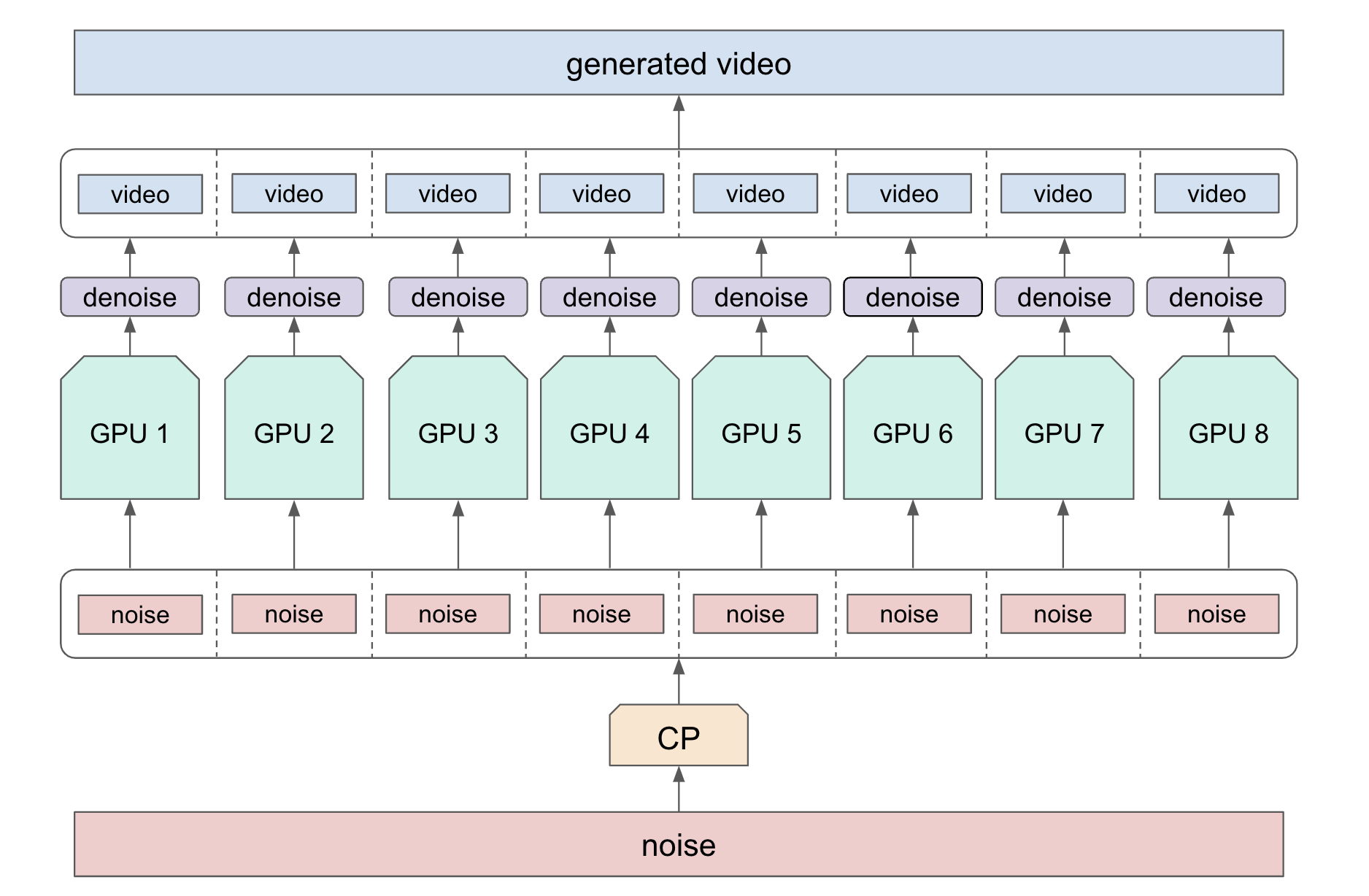}
    \caption{\textbf{Parallelized Inference with Context Parallel.} To generate high-resolution videos efficiently, we leverage CP to parallelize the diffusion denoising process on multiple GPUs. This strategy provides near-linear scaling in inference speedup with respect to the number of GPUs.}
    \label{fig:cp_inference}
\end{figure}

We visualize the process to generate a single video in Figure~\ref{fig:cp_inference}. We begin the video generation process by partitioning the input noise latent representations along the sequence dimension and allocate each chunk to a specific GPU. Once the input noise latents are distributed, the DiT on each GPU independently performs the denoising process on its assigned subsequence. This process is repeated $T$ times, where $T$ is the total number of denoising steps. The denoising operation is a core component of diffusion models, involving iterative refinement to produce coherent and high-quality outputs. By executing these operations in parallel across GPUs, we effectively reduce the overall computation time proportional to the number of GPUs. Finally, the denoised latent tensors are gathered and concatenated to reconstruct the full sequence of the video in latent space. The combined latent sequence is then decoded with the Cosmos Video Tokenizer~\citep{nvidia2025cosmosworldfoundationmodel}.

\subsection{Performance Study}

We benchmark inference performance of our video diffusion model across a variety of configurations. Specifically, we analyze the performance of the NVIDIA Cosmos-1.0-Diffusion-7B-Text2World model~\citep{nvidia2025cosmosworldfoundationmodel}. We use classifier-free guidance (CFG)~\citep{ho2022classifierfreediffusionguidance} during inference, which requires both a conditional and unconditional output simultaneously. Thus, we use a global and micro batch size of 2 during all experiments. We observe that using context parallelism and tensor parallelism at inference has 80-90\% scaling efficiency up to 32 H100 GPUs. We can further improve the speedup by enabling compute/communication overlap for context parallelism to achieve near linear scaling. We also note that on H100 GPUs, FP8 Multi-Head Attention (MHA) from TransformerEngine can improve performance by $\sim$28\% over BF16 on 1 GPU and $\sim$48\% on 32 GPUs. Our inference pipeline enables fast video generation with large models even across multi-node systems, which is crucial to serve these models to customers and utilize them for synthetic data generation pipelines for physical AI applications. There are several other strategies that can also improve inference performance, including model quantization~\citep{li2023qdiffusionquantizingdiffusionmodels}, CFG parallel~\citep{fang2024xdit}, and model distillation~\citep{xie2024emdistillationonestepdiffusion, zhou2024simplefastdistillationdiffusion}. We leave these explorations to future works.

\begin{figure}[htbp]
    \centering
    \includegraphics[width=0.82\linewidth]{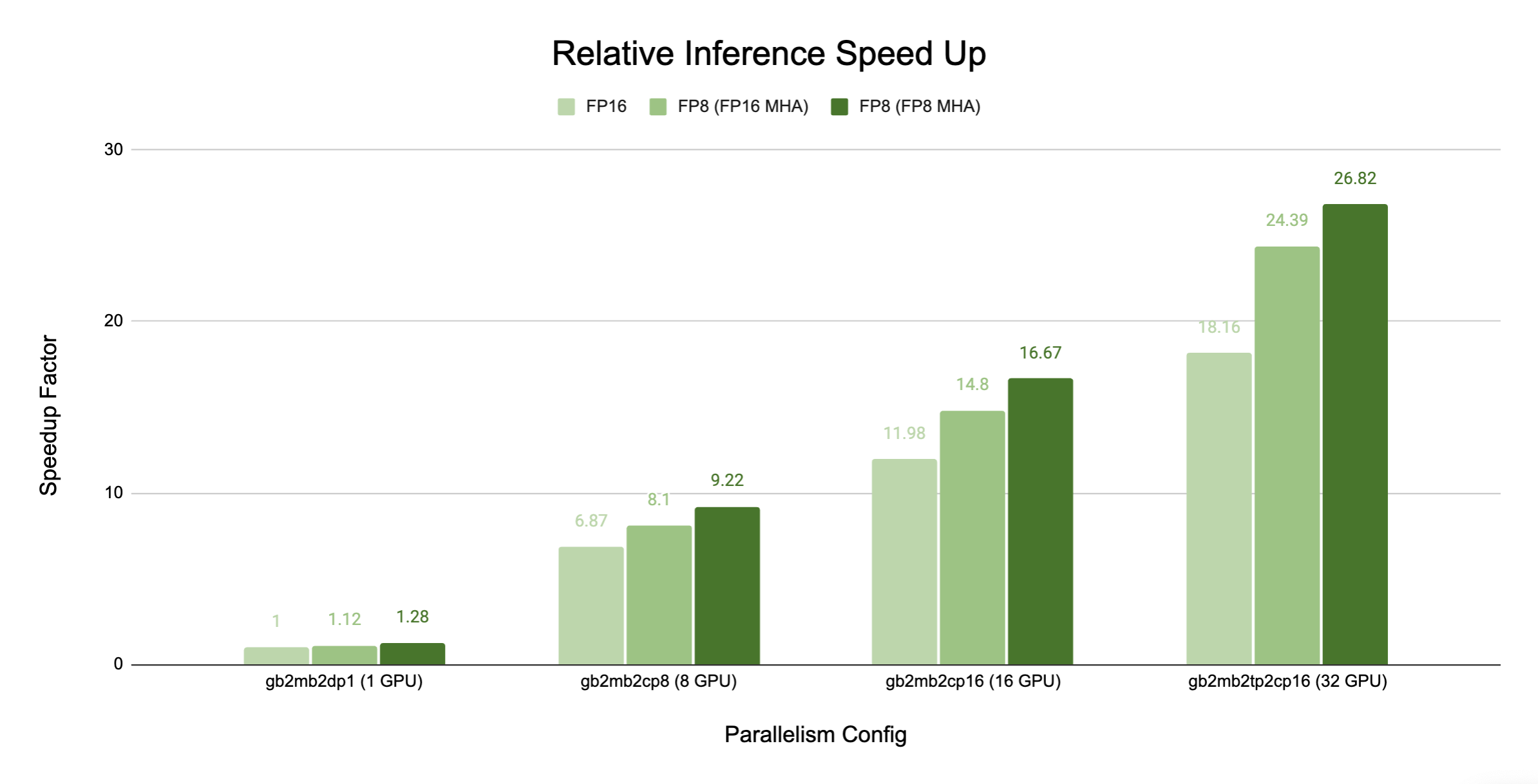}
    \caption{\textbf{Inference Performance Analysis.} We benchmark the Cosmos-1.0-Diffusion-7B-Text2World model across different compute, precision, and model parallelism settings. Our experiments show that we can achieve near-linear scaling with our proposed CP-accelerated inference algorithm.}
    \label{fig:inference_performance}
\end{figure}

%% file: sections/conclusion.tex
\section{Conclusion}
In this paper, we introduced a comprehensive and scalable framework for training VFMs using NVIDIA NeMo. Our framework integrates powerful video curation capabilities via NeMo Curator, multimodal dataloading with Megatron Energon, and scalable video diffusion model training with Megatron Core. We demonstrated the importance of algorithm-system co-design, highlighting the need for advanced 4D parallelization to efficiently train large diffusion transformers. We applied and showed the benefits of algorithm-system co-design in our new hybrid parallelism approach for custom diffusion transformers. Our extensive benchmarking demonstrated superior performance and scalability compared to existing methods at both training and inference time, achieving near-linear scaling and significantly higher MFU than previous baselines. By systematically exploring various parallelism strategies, we established guidelines for optimal parallel configurations across different model sizes and sequence lengths. Our flexible framework sets a strong foundation by providing users the tools needed to explore and scale video foundation models to accelerate novel research and applications.

%% file: sections/contribution.tex
\section{Contributors}\label{sec:contributors}
\subsection{Core Contributors}
\begin{itemize}
    \setlength\itemsep{8pt}
    \item \textbf{NeMo Curator}
    \\
    Ryan Wolf, Niket Agarwal, Jacob Huffman
    \item \textbf{Multimodal Dataloading}
    \\
    Ethan He, Zeeshan Patel
    \item \textbf{Tokenizer Training Pipeline}
    \\
    Linnan Wang, Ethan He
    \item \textbf{Diffusion Training Pipeline}
    \\
    Zeeshan Patel, Ethan He, Parth Mannan, Jack Chang
    \item \textbf{DiT Performance Study}
    \\
    Parth Mannan, Yan Bai, Zeeshan Patel, Ethan He
    \item \textbf{ST-DiT Hybrid Parallel Algorithm Design}
    \\
    Zeeshan Patel, Ethan He, Xiaowei Ren
    \item \textbf{ST-DiT Hybrid Parallel Implementation \& Benchmarking}
    \\
    Zhuoyao Wang, Carl Wang
    \item \textbf{Inference Performance Study}
    \\
    Tommy Huang, Zeeshan Patel, Ethan He
\end{itemize}
\subsection{Contributors}
Sahil Jain, Shanmugam Ramasamy, Joseph Jennings, Ekaterina Sirazitdinova, Oleg Sudakov, Mingyuan Ma, Bobby Chen, Forrest Lin, Hao Wang, Vasanth Rao Naik Sabavat, Sriharsha Niverty, Rong Ou, Pallab Bhattacharya, David Page, Nima Tajbakhsh, Ashwath Aithal
\subsection{Acknowledgements}
We thank Ming-Yu Liu, Yen-Chen Lin, Qinsheng Zhang, and Wenwen Gao for their help and support throughout this project.